\documentclass[11pt,a4paper]{article}
\usepackage[hyperref]{emnlp2020}
\usepackage{times}
\usepackage{latexsym}
\usepackage{microtype}
\usepackage{graphicx}

\usepackage{times}
\usepackage{latexsym}
\usepackage{graphicx}
\usepackage{xspace}
\usepackage{enumitem}
\usepackage{hyperref}
\usepackage{url}
\usepackage{fnpct}
\usepackage{comment}
\usepackage{pifont}
\usepackage{amsmath}
\usepackage{pgfplots}
\pgfplotsset{compat=1.16}

\DeclareMathOperator{\sigmoid}{sigmoid}
\DeclareMathAlphabet{\mathbcal}{OMS}{cmsy}{b}{n}

\newcommand{\mlp}{\textsc{mlp}\xspace}

\newcommand{\bilstm}{\textsc{bilstm}\xspace}
\newcommand{\bigru}{\textsc{bigru}\xspace}
\newcommand{\gru}{\textsc{gru}\xspace}
\newcommand{\wordvec}{\textsc{word2vec}\xspace}
\newcommand{\nodevec}{\textsc{node2vec}\xspace}
\newcommand{\rnn}{\textsc{rnn}\xspace}
\newcommand{\cnn}{\textsc{cnn}\xspace}
\newcommand{\gcn}{\textsc{gcn}\xspace}

\newcommand{\lstm}{\textsc{lstm}\xspace}

\newcommand{\xmtc}{\textsc{xmtc}\xspace}
\newcommand{\lmtc}{\textsc{lmtc}\xspace}

\newcommand{\lwan}{\textsc{lwan}\xspace}

\newcommand{\lwancnn}{\textsc{cnn-lwan}\xspace}

\newcommand{\lwangru}{\textsc{bigru-lwan}\xspace}
\newcommand{\lwangruelmo}{\textsc{bigru-lwan-elmo}\xspace}
\newcommand{\clwangru}{\textsc{c-bigru-lwan}\xspace}
\newcommand{\dclwangru}{\textsc{dc-bigru-lwan}\xspace}

\newcommand{\dnlwangru}{\textsc{dn-bigru-lwan}\xspace}
\newcommand{\dnclwangru}{\textsc{dnc-bigru-lwan}\xspace}
\newcommand{\gclwangru}{\textsc{gc-bigru-lwan}\xspace}
\newcommand{\gnclwangru}{\textsc{gnc-bigru-lwan}\xspace}
\newcommand{\attentionxml}{\textsc{attention-xml}\xspace}
\newcommand{\lwanbertbase}{\textsc{bert-base-lwan}\xspace}
\newcommand{\lwanbert}{\textsc{bert-lwan}\xspace}
\newcommand{\roberta}{\textsc{roberta}\xspace}
\newcommand{\robertabase}{\textsc{roberta-base}\xspace}

\newcommand{\glove}{\textsc{glove}\xspace}
\newcommand{\elmo}{\textsc{elmo}\xspace}
\newcommand{\bert}{\textsc{bert}\xspace}

\newcommand{\bertbase}{\textsc{bert-base}\xspace}
\newcommand{\tfive}{\textsc{t5-3b}\xspace}
\newcommand{\megatronlm}{\textsc{megatron-lm}\xspace}

\newcommand{\clibert}{\textsc{clinical-bert}\xspace}
\newcommand{\scibert}{\textsc{sci-bert}\xspace}
\newcommand{\hierscibert}{\textsc{hier-sci-bert}\xspace}

\newcommand{\eurlexdata}{\textsc{eurlex57k}\xspace}
\newcommand{\amazondata}{\textsc{amazon13k}\xspace}
\newcommand{\amazon}{\textsc{amazon}\xspace}
\newcommand{\plts}{\textsc{plt}s\xspace}
\newcommand{\plt}{\textsc{plt}\xspace}
\newcommand{\bpe}{\textsc{bpe}\xspace}
\newcommand{\cls}{\textsc{cls}\xspace}
\newcommand{\eurovoc}{\textsc{eurovoc}\xspace}

\newcommand{\icdix}{\textsc{icd-9}\xspace}

\newcommand{\eurlex}{\textsc{eur-lex}\xspace}

\newcommand{\nlp}{\textsc{nlp}\xspace}

\newcommand{\mimiciii}{\textsc{mimic-iii}\xspace}
\newcommand{\tfidf}{\textsc{tf-idf}\xspace}

\newcommand{\parabel}{\textsc{parabel}\xspace}
\newcommand{\bonsai}{\textsc{bonsai}\xspace}

\newcommand\gap{\textsc{gap}\xspace}
\newcommand{\gclwancnn}{\textsc{gc-cnn-lwan}\xspace}
\newcommand{\clwancnn}{\textsc{c-cnn-lwan}\xspace}

\aclfinalcopy

\title{An Empirical Study on Large-Scale Multi-Label Text Classification Including Few and Zero-Shot Labels}

\author{Ilias Chalkidis$^{\;\dagger\;\ddagger}$ \qquad Manos Fergadiotis$^{\;\dagger\;\ddagger}$ \qquad Sotiris Kotitsas$^{\;\dagger}$ \\ \textbf{Prodromos Malakasiotis$^{\;\dagger\;\ddagger}$} \qquad \textbf{Nikolaos Aletras$\;^{*}$} \qquad \textbf{Ion Androutsopoulos$^{\;\dagger\;\ddagger}$} \\$^{\dagger\;}$Department of Informatics, Athens University of Economics and Business \\ $^{\ddagger\;}$Institute of Informatics \& Telecommunications, NCSR ``Demokritos'' \\ $^{*\;}$Computer Science Department, University of Sheffield, UK \\ {\tt [ihalk,fergadiotis,kotitsas,rulller,ion]@aueb.gr} \\ {\tt n.aletras@sheffield.ac.uk}}

\date{}

\begin{document}
\maketitle

\begin{abstract}
Large-scale Multi-label Text Classification (\lmtc) has a wide range of Natural Language Processing (\nlp) applications and presents interesting challenges. First, not all labels are well represented in the training set, due to the very large label set and the skewed label distributions of \lmtc datasets. Also, label hierarchies and differences in human labelling guidelines may affect graph-aware annotation proximity. Finally, the label hierarchies are periodically updated, requiring \lmtc models capable of zero-shot generalization. Current state-of-the-art \lmtc models employ Label-Wise Attention Networks (\lwan{s}), which (1) typically treat \lmtc as flat multi-label classification; (2) may use the label hierarchy to improve zero-shot learning, although this practice is vastly understudied; and (3) have not been combined with pre-trained Transformers (e.g.\ \bert), which have led to state-of-the-art results in several \nlp benchmarks. Here, for the first time, we empirically evaluate a battery of \lmtc methods from vanilla \lwan{s} to hierarchical classification approaches and transfer learning, on frequent, few, and zero-shot learning on three datasets from different domains. We show that hierarchical methods based on Probabilistic Label Trees (\plt{s}) outperform \lwan{s}. Furthermore, we show that Transformer-based approaches outperform the state-of-the-art in two of the datasets, and we propose a new state-of-the-art method which combines \bert with \lwan. Finally, we propose new models that leverage the label hierarchy to improve few and zero-shot learning, considering on each dataset a graph-aware annotation proximity measure that we introduce.
\end{abstract}

\section{Introduction}
Large-scale Multi-label Text Classification (\lmtc) is the task of assigning a subset of labels from a large predefined set (typically thousands) to a given document. \lmtc has a wide range of applications in Natural Language Processing (\nlp), such as associating medical records with diagnostic and procedure labels \cite{Mullenbach2018,Rios2018-2}, legislation with relevant legal concepts \cite{Mencia2007, Chalkidis2019}, and products with categories \cite{Lewis2004,Partalas2015LSHTCAB}.

\begin{figure}[t]
\centering
\includegraphics[width=\columnwidth]{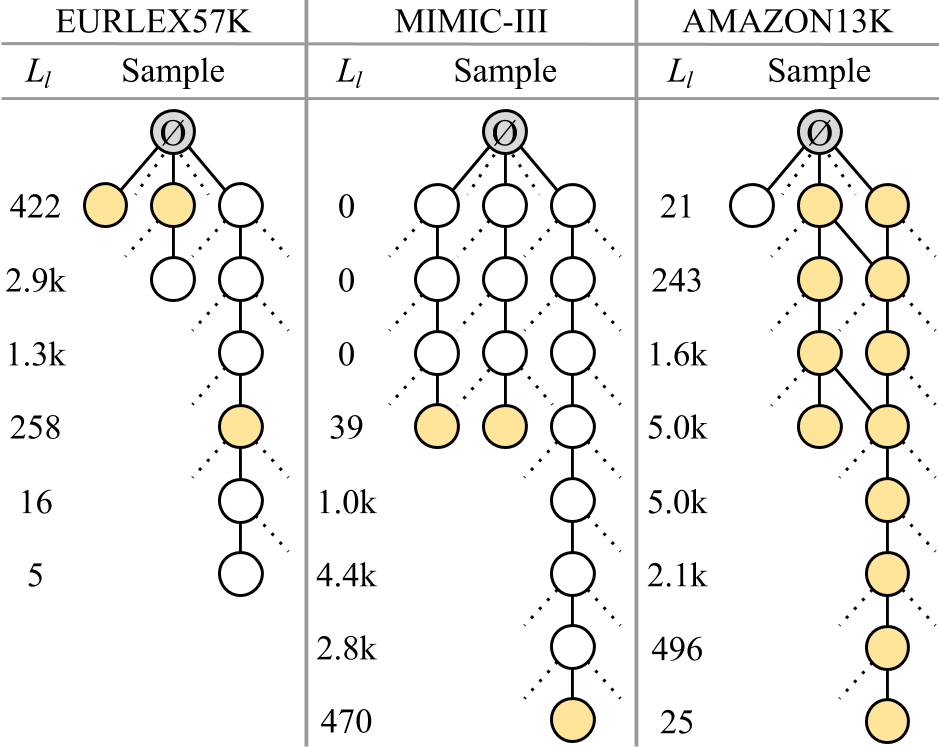}
\caption{Examples from \lmtc label hierarchies. $\emptyset$ is the root label. $L_l$ is the number of labels per level. Yellow nodes denote gold label assignments. In \eurlexdata, documents have been tagged with both leaves and inner nodes (\gap: 0.45). In \mimiciii, only leaf nodes can be used, causing the label assignments to be much sparser (\gap: 0.27). In \amazondata, documents are tagged with leaf nodes, but it is assumed that all the parent nodes are also assigned, leading to dense label assignments (\gap: 0.86).}
\label{fig:hierarchies}
\vspace{-5mm}
\end{figure}

Apart from the large label space, \lmtc datasets often have skewed label distributions (e.g., some labels have few or no training examples) and a label hierarchy with different labelling guidelines (e.g., they may require documents to be tagged only with leaf nodes, or they may allow both leaf and other nodes to be used). The latter affects graph-aware annotation proximity (\gap), i.e., the proximity of the gold labels in the label hierarchy (see Section~\ref{sec:density}). Moreover, the label set and the hierarchies are periodically updated, thus requiring zero- and few-shot learning to cope with newly introduced labels. Figure~\ref{fig:hierarchies} shows a sample of label hierarchies, with different label assignment guidelines, from three standard \lmtc benchmark datasets: \eurlex \cite{Chalkidis2019}, \mimiciii \cite{Johnson2017}, and \amazon \cite{McAuley2013}). 

Current state-of-the-art \lmtc models are based on Label-Wise Attention Networks (\lwan{s}) \citep{Mullenbach2018}, which use a different attention head for each label. \lwan{s} (1) typically do not leverage structural information from the label hierarchy, treating \lmtc as flat multi-label classification; (2) may use the label hierarchy to improve performance in few/zero-shot scenario, but this practice is vastly understudied; and (3) have not been combined with pre-trained Transformers. 

We empirically evaluate, for the first time, a battery of \lmtc methods, from vanilla \lwan{s} to  hierarchical classification approaches and transfer learning, in frequent, few, and zero-shot learning scenarios. We experiment with three standard \lmtc datasets (\eurlexdata; \mimiciii; \amazondata). Our contributions are the following: 
\begin{itemize}
    \item We show that hierarchical \lmtc approaches based on Probabilistic Label Trees (\plt{s})~\cite{Prabhu2018,Khandagale2019,You2019} outperform flat neural state-of-the-art methods, i.e., \lwan \cite{Mullenbach2018} in two out of three datasets (\eurlexdata, \amazondata).
     \item We demonstrate that pre-trained Transformer-based approaches (e.g., \bert) further improve the results in two of the three datasets (\eurlexdata, \amazondata), and we propose a new method that combines \bert with \lwan achieving the best results overall. 
    \item Finally, following the work of \citet{Rios2018-2} for few and zero-shot learning on \mimiciii, we investigate the use of structural information from the label hierarchy in \lwan. We propose new \lwan-based models with improved performance in these settings, taking into account the labelling guidelines of each dataset and a graph-aware annotation proximity (\gap) measure that we introduce.
\end{itemize}

\section{Related Work}
\label{sec:related_work}

\subsection{Advances and limitations in LMTC}
In \lmtc, deep learning achieves state-of-the-art results with \lwan{s} \citep{You2018,Mullenbach2018,Chalkidis2019}, in most cases comparing to naive baselines (e.g., vanilla \cnn{s} or vanilla \lstm{s}). The computational complexity of \lwan{s}, however, makes it difficult to scale them up to extremely large label sets. Thus, Probabilistic Label Trees (\plts) \cite{Jasinska2016,Prabhu2018,Khandagale2019} are preferred in Extreme Multi-label Text Classification (\xmtc), mainly because the linear classifiers they use at each node of the partition trees can be trained independently considering few labels at each node. This allows \plt{-based} methods to efficiently handle extremely large label sets (often millions), while also achieving top results in \xmtc. Nonetheless, previous work has not thoroughly compared \plt-based methods to neural models in \lmtc. In particular, only \citet{You2018} have compared \plt methods to neural models in \lmtc, but without adequately tuning their parameters, nor considering few and zero-shot labels. More recently, \citet{You2019} introduced \attentionxml, a new method primarily intended for \xmtc, which combines \plt{s} with \lwan classifiers. Similarly to the rest of \plt{-based} methods, it has not been evaluated in \lmtc.

\subsection{The new paradigm of transfer learning}
Transfer learning \cite{ruder2019,Rogers2020API}, which has recently achieved state-of-the-art results in several \nlp tasks, has only been considered in legal \lmtc by \newcite{Chalkidis2019}, who experimented with \bert \cite{bert} and \elmo \cite{Peters2018}. Other \bert variants, e.g.\ \roberta \cite{roberta}, or \bert-based models have not been explored in \lmtc so far.

\subsection{Few and zero-shot learning in LMTC}
Finally, few and zero-shot learning in \lmtc is mostly understudied. \citet{Rios2018-2} investigated the effect of encoding the hierarchy in these settings, with promising results. However, they did not consider other confounding factors, such as using deeper neural networks at the same time, or alternative encodings of the hierarchy. \citet{Chalkidis2019} also considered few and zero-shot learning, but ignoring the label hierarchy. 

\vspace{2mm}\noindent Our work is the first attempt to systematically compare flat, \plt{-based}, and hierarchy-aware \lmtc methods in frequent, few-, and zero-shot learning, and the first exploration of the effect of transfer learning in \lmtc on multiple datasets.

\section{Models}

\subsection{Notation for neural methods}
We experiment with neural methods consisting of: (i) a \emph{token encoder} ($\mathcal{E}_w$), which makes token embeddings ($w_{t}$) context-aware ($h_{t}$); (ii) a \emph{document encoder} ($\mathcal{E}_d$), which turns a document into a single embedding; (iii) an optional \emph{label encoder} ($\mathcal{E}_l$), which turns each label into a label embedding; (iv) a \emph{document decoder} ($\mathcal{D}_d$), which maps the document to label probabilities. Unless otherwise stated, tokens are words, and $\mathcal{E}_w$ is a stacked \bigru. 

\subsection{Flat neural methods}
\noindent\textbf{\lwangru:} In this model \cite{Chalkidis2019}\footnote{The original model was proposed by \citet{Mullenbach2018}, with a \cnn token encoder ($\mathcal{E}_w$). \newcite{Chalkidis2019} show that \bigru is a better encoder than \cnn{s}. See also the supplementary material for a detailed comparison.}, $\mathcal{E}_d$ uses one  attention head per label to generate $L$ document representations $d_l$:
\vspace{-1mm}
\begin{flalign}
a_{lt} = \frac{\mathrm{exp}(h_t^\top u_{l})}{\sum_{t'} \mathrm{exp}(h_{t'}^\top u_{l})}, \;
d_l =  \frac{1}{T} \sum^T_{t=1} a_{lt} h_t
\end{flalign}

\noindent $T$ is the document length in tokens, $h_t$ the context-aware representation of the $t$-th token, and $u_l$ a trainable vector used to compute the attention scores of the $l$-th attention head; $u_l$ can also be viewed as a label representation. Intuitively, each head focuses on possibly different tokens of the document to decide if the corresponding label should be assigned. In this model, $\mathcal{D}_d$ employs $L$ linear layers with $\sigmoid$ activations, each operating on a different label-wise document representation $d_l$, to produce the probability of the corresponding label.

\subsection{Hierarchical PLT-based methods} 
In \plt{-based} methods, each label is represented as the average of the feature vectors of the training documents that are annotated with this label. The root of the \plt corresponds to the full label set. The label set is partitioned into $k$ subsets using $k$-means clustering, and each subset is represented by a child node of the root in the \plt. The labels of each new node are then recursively partitioned into $k$ subsets, which become children of that node in the \plt. If the label set of a node has fewer than $m$ labels, the node becomes a leaf and the recursion terminates. During inference, the \plt is traversed top down. At each non-leaf node, a multi-label classifier decides which children nodes (if any) should be visited by considering the feature vector of the document.
When a leaf node is visited, the multi-label classifier of that node decides which labels of the node will be assigned to the document.
\vspace{1mm} 

\noindent\textbf{\parabel, \bonsai:} We experiment with \parabel \cite{Prabhu2018} and \bonsai \cite{Khandagale2019}, two state-of-the-art \plt-based methods. \parabel employs binary  \plt{s} ($k=2$), while \bonsai uses non-binary \plt{s} ($k>2$), which are shallower and wider. 
In both methods, a linear classifier is used at each node, and documents are represented by \tfidf feature vectors.
\vspace{2mm}

\noindent\textbf{\attentionxml:} Recently, \citet{You2019} proposed a hybrid method that aims to leverage the advantages of both \plt{s} and neural models. Similarly to \bonsai, \attentionxml uses non-binary trees. However, the classifier at each node of the \plt is now an \lwan with a \bilstm token encoder ($\mathcal{E}_w$), instead of a linear classifier operating on \tfidf document representations.

\subsection{Transfer learning based LMTC}
\noindent\textbf{\lwangruelmo:}
In this model, we use \elmo \cite{Peters2018} to obtain context-sensitive token embeddings, which we concatenate with the pre-trained word embeddings to obtain the initial token embeddings ($w_t$) of \lwangru. Otherwise, the model is the same as \lwangru.
\vspace{2mm} 

\noindent\textbf{\bert, \roberta:} Following \citet{bert}, we feed each document to \bert and obtain the top-level representation $h_{\text{\cls}}$ of \bert{'s} special \cls token as the (single) document representation. $\mathcal{D}_d$ is now a linear layer with $L$ outputs and $\sigmoid$ activations which operates directly on $h_{\text{\cls}}$, producing a probability for each label. The same arrangement applies to \roberta \cite{roberta}.\footnote{Unlike \bert, \roberta uses dynamic masking, it eliminates the next sentence prediction pre-training task, and uses a larger vocabulary. \newcite{roberta} reported better results in \nlp benchmarks using \roberta.} 
\vspace{2mm} 

\noindent\textbf{\lwanbert:}  Given the large size of the label set in \lmtc datasets, we propose a combination of \bert and \lwan. Instead of using $h_{\text{\cls}}$ as the document representation and pass it through a linear layer with $L$ outputs (as with \bert and \roberta), we pass all the top-level output representations of \bert into a label-wise attention mechanism. The entire model (\lwanbert) is jointly trained, also fine-tuning the underlying \bert encoder.

\subsection{Zero-shot LMTC}
\label{sec:zero-shot)}
\noindent\textbf{\clwangru} is a zero-shot capable extension of \lwangru. It was proposed by \citet{Rios2018-2}, but with a \cnn encoder; instead, we use a \bigru. In this method, $\mathcal{E}_l$ creates $u_l$ as the \emph{centroid} of the token embeddings of the corresponding label descriptor.
The label representations $u_l$ are then used by the attention heads.\vspace{-1mm}
\begin{gather}
v_t  = \tanh(W h_t + b) \label{eq:lw_project}\\
a_{lt}  = \frac{\mathrm{exp}(v_t^\top u_{l})}{\sum_{t'} \mathrm{exp}(v_{t'}^\top u_{l})}, 
\;\;
d_l  =  \frac{1}{T} \sum^T_{t=1} a_{lt} h_t \label{eq:lw_doc}
\end{gather}
Here $h_t$ are the context-aware embeddings of $\mathcal{E}_w$, $a_{lt}$ is the attention score of the $l$-th attention head for the $t$-th document token, viewed as $v_t$ (Eq.\ 2), $d_l$ is the label-wise document representation for the $l$-th label.
$\mathcal{D}_d$ also relies on the label representations $u_l$ to produce each label probability $p_l$.\vspace{-1mm}
\begin{flalign}
p_l = \sigmoid(u_l^\top d_l) 
\label{eq:p_l}
\end{flalign}
The centroid label representations $u_l$ of both encountered (during training) and unseen (zero-shot) labels remain unchanged, because the token embeddings in the centroids are not updated. This keeps the representations of unseen labels close to those of similar labels encountered during training. 
In turn, this helps the attention mechanism (Eq.~\ref{eq:lw_doc}) and the decoder (Eq.~\ref{eq:p_l}) cope with unseen labels that have similar descriptors with encountered labels.\vspace{2mm}

\noindent\textbf{\gclwangru:} This model, originally proposed by \citet{Rios2018-2}, applies graph convolutions (\gcn{s}) to the label hierarchy.\footnote{The original model uses a \cnn token encoder ($\mathcal{E}_w$), whereas we use a \bigru encoder, which is a better encoder. See the supplementary material for a detailed comparison.} The intuition is that the \gcn{s} will help the representations of rare labels benefit from the (better) representations of more frequent labels that are nearby in the label hierarchy. $\mathcal{E}_l$ now creates graph-aware label representations $u^{3}_l$ from the corresponding label descriptors (we omit the bias terms for brevity) as follows:
\vspace*{-1mm}
\begin{flalign}
u^{1}_l\! &=\! f(W^{1}_s u_l +\!\! \sum_{j\in N_{p,l}}\! \frac{W^{1}_p u_j}{\left|N_{p,l}\right|} +\!\! \sum_{j\in N_{c,l}}\! \frac{W^{1}_c u_j}{\left|N_{c,l}\right|}) \label{eq:graph1} \\
u^{2}_l\! &=\! f(W^{2}_s u^1_l +\!\! \sum_{j\in N_{p,l}}\! \frac{W^{2}_p u^1_j}{\left|N_{p,l}\right|} +\!\! \sum_{j\in N_{c,l}}\! \frac{W^{2}_c u^1_j}{\left|N_{c,l}\right|}) \label{eq:graph2} \\
u^{3}_l\! &=\! [ u_l ; u^{2}_l ] \label{eq:graph3}
\end{flalign}
where $u_l$ is again the centroid of the token embeddings of the descriptor of the $l$-th label; $W^{i}_s$, $W^{i}_p$, $W^{i}_c$ are matrices for self, parent, and children nodes of each label; $N_{p,l}$, $N_{c,l}$ are the sets of parents and children of the the $l$-th label; and $f$ is the $\tanh$ activation. The label-wise document representations $d_l$ are again produced by $\mathcal{E}_d$, as in \clwangru (Eq.~\ref{eq:lw_project}--\ref{eq:lw_doc}), but they go through an additional dense layer with $\tanh$ activation (Eq.~\ref{eq:doc_project_graph}). The resulting document representations $d_{l,o}$ and the graph-aware label representations $u^{3}_l$ are then used by $\mathcal{D}_d$ to produce a  probability $p_l$ for each label (Eq.~\ref{eq:pl_graph}).
\vspace*{-1mm}
\begin{flalign}
d_{l,o} &= \mathrm{tanh}(W_o d_l + b_o) \label{eq:doc_project_graph}\\
p_l &=  \sigmoid\left((u^3_l)^\top d_{lo} \label{eq:pl_graph}\right)
\end{flalign}

\noindent\textbf{\dclwangru:} The stack of \gcn layers in \gclwangru (Eq.~\ref{eq:graph1}--\ref{eq:graph2}) can be turned into a plain two-layer Multi-Layer Perceptron (\mlp), unaware of the label hierarchy, by setting $N_{p,l} = N_{c,l} = \emptyset$. We call \dclwangru the resulting (deeper than \clwangru) variant of \gclwangru. We use it as an ablation method to evaluate the impact of the \gcn layers on performance.\vspace{2mm}

\noindent\textbf{\dnlwangru:} As an alternative approach to exploit the label hierarchy, we used a recent improvement of \nodevec \cite{grover2016} by \newcite{kotitsas2019} to obtain alternative hierarchy-aware label representations. \nodevec  is similar to \wordvec \cite{Mikolov13}, but pre-trains node embeddings instead of word embeddings, replacing \wordvec's text windows by random walks on a graph (here the label hierarchy).\footnote{The \nodevec extension we used also considers the textual descriptors of the nodes, using an \rnn encoder operating on token embeddings.} 
In a variant of \dclwangru, dubbed \dnlwangru, we simply replace the initial centroid $u_l$ label representations of \dclwangru in Eq.~\ref{eq:graph1} and \ref{eq:graph3} by the label representations $g_l$ generated by the \nodevec extension.\vspace{2mm}

\noindent\textbf{\dnclwangru:} In another version of \dclwangru, called \dnclwangru, we replace the initial centroid $u_l$ label representations of \dclwangru by the concatenation $[u_l;g_l]$.\vspace{2mm}

\noindent\textbf{\gnclwangru:} Similarly, we expand \gclwangru with the hierarchy-aware label representations of the \nodevec extension. Again, we replace the  centroid $u_l$ label representations of \gclwangru in Eq.~\ref{eq:graph1} and \ref{eq:graph3} by the label representations $g_l$ of the \nodevec extension. The resulting model, \gnclwangru, uses both \nodevec and the \gcn layers to encode the label hierarchy, thus obtaining knowledge from the label hierarchy both in a self-supervised and a supervised fashion.

\section{Experimental Setup}
\label{experiments}

\subsection{Graph-aware Annotation Proximity}
\label{sec:density}

In this work, we introduce \emph{graph-aware label proximity} (\gap), a measure of the topological proximity (on the label hierarchy) of the gold labels assigned to documents. \gap turns out to be a key factor in the performance of hierarchy-aware zero-shot capable extensions of \lwangru. Let $G(L, E)$ be the graph of the label hierarchy, where $L$ is the set of nodes (label set) and $E$ the set of edges. Let $L_d \subseteq L$ be the set of gold labels a particular document $d$ is annotated with. Finally, let $G_d^+(L_d^+, E_d^+)$ be the minimal (in terms of $|L_d^+|$) subgraph of $G(L, E)$, with  $L_d \subseteq L^{+}_d \subseteq L$ and $E^{+}_d \subseteq E$, such that for any two nodes (gold labels) $l_1, l_2 \in L_d$, the shortest path between $l_1, l_2$ in the full graph $G(L,E)$ is also a path in $G_d^+(L_d^+, E_d^+)$. Intuitively, we extend $L_d$ to $L_d^+$ by including additional labels that lie between any two assigned labels $l_1, l_2$ on the shortest path that connects $l_1, l_2$ in the full graph. We then define $\gap_d=\frac{|L_d|}{|L^{+}_d|}$. By averaging $\gap_d$ over all the documents $d$ of a dataset, we obtain a single \gap score per dataset (Fig.~\ref{fig:hierarchies}). When the assigned (gold) labels of the documents are frequently neighbours in the full graph (label hierarchy), we need to add fewer labels when expanding the $L_d$ of each document to $L_d^+$; hence, $\gap \rightarrow 1$. When the assigned (gold) labels are frequently remote to each other, we need to add more labels ($|L_d^+| \gg |L_d|$) and $\gap \rightarrow 0$. 

\gap should not be confused with \emph{label density} \cite{tsoumakas2009}, defined as $\mathrm{D}={\frac {1}{N}}\sum^N_{d=1}{\frac{|L_{d}|}{|L|}}$, where $N$ is the total number of documents. Although label density is often used in the multi-label classification literature, it is graph-unaware, i.e., it does not consider the positions (and distances) of the assigned labels  in the graph.

\subsection{Data}
\label{sec:datasets}

\textbf{\eurlexdata}
\citep{Chalkidis2019} contains 57k English legislative documents from \eurlex.\footnote{\url{http://eur-lex.europa.eu/}}
Each document is annotated with one or more concepts (labels) from the  4,271 concepts of \eurovoc.\footnote{\url{http://eurovoc.europa.eu/}} The average document length is approx.\ 727 words.
The labels are divided in \emph{frequent} (746 labels), \emph{few-shot} (3,362), and \emph{zero-shot} (163), depending on whether they were assigned to $n > 50$, $1 < n \leq 50$, or no training documents.
They are organized in a 6-level hierarchy, which was not considered in the experiments of \citet{Chalkidis2019}. The documents are labeled with concepts from all levels (Fig.~\ref{fig:hierarchies}), but in practice if a label is assigned, none of its ancestor or descendent labels are assigned. The resulting \gap is 0.45.
\vspace{2mm}

\noindent\textbf{\mimiciii} \cite{Johnson2017} contains approx.\ 52k English discharge summaries from \textsc{us} hospitals. The average document length is approx.\ 1.6k words. Each summary has one or more codes (labels) from 8,771 leaves of the \icdix hierarchy, which has 8 levels (Fig.~\ref{fig:hierarchies}).\footnote{\url{www.who.int/classifications/icd/en/}} Labels are divided in \emph{frequent} (4,112 labels), \emph{few-shot} (4,216 labels), and \emph{zero-shot} (443 labels), depending on whether they were assigned to $n>5$, $1< n \leq 5$, or no training documents.  All discharge summaries are annotated with leaf nodes (5-digit codes) only, i.e., the most fine-grained categories (Fig.~\ref{fig:hierarchies}), causing the label assignments to be much sparser compared to \eurlexdata (\gap  0.27).\vspace{2mm}

\noindent\textbf{\amazondata} \cite{Lewis2004} contains approx.\ 1.5M English product descriptions from Amazon. Each product is represented by a title and a description, which are on average 250 words when concatenated. 
Products are classified into one or more categories (labels) from a set of approx.\ 14k. Labels are divided in \emph{frequent} (3,108 labels), \emph{few-shot} (10,581 labels), \emph{zero-shot} (579 labels), depending on whether they were assigned to $n > 100$, $1 < n \leq 100$, or no training documents. The labels are organized in a hierarchy of 8 levels. If a product is annotated with a label, all of its ancestor labels are also assigned to the product (Fig.~\ref{fig:hierarchies}), leading to dense label assignments (\gap 0.86).

\begin{table*}[ht!]
{\footnotesize
\centering
\resizebox{\textwidth}{!}{
\begin{tabular}{lcc|cc|cc}
  \hline
   & \multicolumn{2}{c}{\textsc{All Labels}} & \multicolumn{2}{c}{\textsc{Frequent}} & \multicolumn{2}{c}{\textsc{Few}} \\
  & \textit{RP@K} & 
  \textit{nDCG@K} & \textit{RP@K} & \textit{nDCG@K} & \textit{RP@K} & \textit{nDCG@K} \\
  \hline
  \hline
  & \multicolumn{6}{c}{\eurlexdata ($L_{AVG}=5.07, K=5$)} \\
  \hline
  \hline
  \textsc{Flat neural methods} & \multicolumn{6}{c}{} \\
  \hline
  \lwangru \cite{Chalkidis2019} & \underline{77.1} & \underline{80.1} & \underline{81.0} & \underline{82.4} & 65.6 & 61.7 \\ 
  \gclwangru \cite{Rios2018-2} & 76.8 & 80.0 & 80.6 & 82.3 & \underline{66.2} & \underline{61.8} \\
  \hline
  \textsc{Hierarchical \textsc{plt}-based methods} & \multicolumn{6}{c}{} \\
  \hline
  \parabel \cite{Prabhu2018} & 78.1 & 80.6 & 82.4 & 83.3 & 59.9 & 57.3 \\ 
  \bonsai \cite{Khandagale2019} & \underline{79.3} & \underline{81.8} & \underline{83.4} & \underline{84.3} & 65.0 & 61.6 \\
  \attentionxml \cite{You2019} & 78.1 & 80.0 & 81.9 & 83.1 & \underline{68.9} & \underline{64.9} \\
 \hline
 \textsc{Transfer learning} & \multicolumn{6}{c}{} \\
  \hline
 \lwangruelmo \cite{Chalkidis2019} & 78.1 & 81.1 & 82.1 & 83.5 & 66.8 & 61.9 \\ 
  \bertbase \cite{bert} & 79.6 & 82.3 & 83.4 & 84.6 & 69.3 & 64.4 \\
  \robertabase \cite{roberta} & 79.3 & 81.9 & 83.4 & 84.4 & 67.5 & 62.4 \\
  \lwanbertbase (new) & \underline{\textbf{80.3}} & \underline{\textbf{82.9}} & \underline{\textbf{84.3}} & \underline{\textbf{85.4}} & \underline{\textbf{69.9}} & \underline{\textbf{65.0}} \\
  \hline
  \hline
 & \multicolumn{6}{c}{\mimiciii ($L_{AVG}=15.45, K=15$)}  \\
  \hline
  \hline
  \textsc{Flat neural methods} & \multicolumn{6}{c}{} \\
  \hline
  \lwangru \cite{Chalkidis2019} & \underline{66.2} & \underline{70.1} & \underline{66.8} & \underline{70.6} & 21.7 & 14.3 \\ 
  \gclwangru \cite{Rios2018-2} & 64.9 & 69.1 & 65.6 & 69.6 & \underline{\textbf{35.9}} & \underline{\textbf{21.1}} \\
    \hline
    \textsc{Hierarchical \textsc{plt}-based methods} & \multicolumn{6}{c}{} \\
    \hline
  \parabel \cite{Prabhu2018} & 58.7 & 63.3 & 59.3 & 63.7 & 9.6 & 6.0 \\ 
  \bonsai \cite{Khandagale2019} & 59.4 & 64.0 & 60.0 & 64.4 & 11.8 & 7.9 \\ 
  \attentionxml \cite{You2019} & \underline{\textbf{69.3}} & \underline{\textbf{73.4}} & \underline{\textbf{70.0}} & \underline{\textbf{73.8}} & \underline{26.9} & \underline{19.5} \\
   \hline
   \textsc{Transfer learning} & \multicolumn{6}{c}{} \\
   \hline
   \lwangruelmo \cite{Chalkidis2019} & \underline{66.8} & \underline{70.9} & \underline{67.5} & \underline{71.3} & \underline{21.2} & \underline{13.0} \\
   \bertbase \cite{bert} & 52.7 & 58.1 & 53.2 & 58.4 & 18.2 & 10.0 \\
  \robertabase \cite{roberta} & 53.7 & 58.9 & 54.3 & 59.2 & 18.1 & 10.9 \\
  \lwanbertbase (new) & 50.1 & 55.2 & 50.6 & 55.5 & 15.3 & 9.1 \\
    \hline
    \hline
  & \multicolumn{6}{c}{\amazondata ($L_{AVG}=5.04, K=5$)}  \\
  \hline
   \hline
   \textsc{Flat neural methods} & \multicolumn{6}{c}{} \\
  \hline
     \lwangru \cite{Chalkidis2019} & \underline{83.9} & \underline{85.4} & \underline{84.9} & \underline{86.1} & \underline{\textbf{80.0}} & \underline{\textbf{73.6}} \\ 
     \gclwangru \cite{Rios2018-2} & 77.4 & 79.8 & 79.1 & 81.0 & 53.7 & 45.8  \\
     \hline
    \textsc{Hierarchical \textsc{plt}-based methods} & \multicolumn{6}{c}{} \\
    \hline
  \parabel \cite{Prabhu2018} & \underline{85.1} & \underline{86.7} & \underline{86.3} & \underline{87.4} & 76.8 & 71.9 \\ 
  \bonsai \cite{Khandagale2019} & \underline{85.1} & 86.6 & 86.2 & 87.3 & \underline{78.3} & \underline{73.2} \\
  \attentionxml \cite{You2019} & 84.9 & 86.7 & 86.0 & \underline{87.4} & 76.0 & 69.7 \\
   \hline
   \hline
   \textsc{Transfer learning} & \multicolumn{6}{c}{} \\
   \hline
  \lwangruelmo \cite{Chalkidis2019} & 
  85.1 & 86.6 & 86.2 & 87.4 & \underline{79.9} & \underline{73.5} \\
  \bertbase \cite{bert} & 86.8 & 88.5 & 88.5 & 89.6 & 70.3 & 62.2 \\
  \robertabase \cite{roberta} & 84.1 & 85.9 & 85.7 & 87.0 & 70.6 & 61.3 \\
  \lwanbertbase (new) & \underline{\textbf{87.3}} & \underline{\textbf{88.9}} & \underline{\textbf{88.8}} & \underline{\textbf{90.0}} & 77.2 & 68.9 \\
  \hline
\end{tabular}
}
}
\caption{Results (\%) of experiments across base methods for all, frequent, and few label groups. All base methods are incapable of zero-shot learning. The best overall results are shown in bold. The best results in each zone are shown underlined. We show results for $K$ close to the average number of labels $L_{AVG}$.}
\vspace*{-5mm}
\label{tab:results}
\end{table*}

\subsection{Evaluation Measures}
The most common evaluation measures in \lmtc are label precision and recall at the top \textit{K} predicted labels (\textit{P@K}, \textit{R@K}) of each document, and \textit{nDCG@K} \cite{Manning2009}, both averaged over test documents. However, \textit{P@K} and \textit{R@K} unfairly penalize methods when the gold labels of a document are fewer or more than \textit{K}, respectively. R-Precision\textit{@K} (\textit{RP@K}) \cite{Chalkidis2019}, a top-\textit{K} version of R-Precision \cite{Manning2009}, is better; it is the same as \textit{P@K} if there are at least \textit{K} gold labels, otherwise \textit{K} is reduced to the number of gold labels. When the order of the top-\textit{K} labels is unimportant (e.g., for small \textit{K}), \textit{RP@K} is more appropriate than \textit{nDCG@K}.

\subsection{Implementation Details}
We implemented neural methods in \textsc{tensorflow 2}, also relying on the  HuggingFace Transformers library for \bert{-based} models.\footnote{Consult \url{https://tersorflow.org/} and \url{http://github.com/huggingface/transformers/}.} We use the \textsc{base} versions of all models, and the Adam optimizer \cite{Kingma2015}.  All hyper-parameters were tuned selecting  values with the best loss on the development data.\footnote{See the appendix for details and hyper-parameters.} For all \plt{-based} methods, we used the code provided by their authors.\footnote{\parabel: \url{http://manikvarma.org/code/Parabel/download.html}; \bonsai: \url{https://github.com/xmc-aalto/bonsai}; \attentionxml: \url{http://github.com/yourh/AttentionXML}}

\section{Results}

\subsection{Overall predictive performance}
\label{sec:discussion}

\noindent\textbf{\plts vs.\ \lwan{s}:} Interestingly, the \tfidf{-based} \parabel and \bonsai outperform the best previously published neural \lwan-based models on \eurlexdata and \amazondata, while being comparable to \attentionxml, when all or frequent labels are considered (Table~\ref{tab:results}). This is not the case with \mimiciii, where \lwangru and \attentionxml have far better results for all and frequent labels. 
The poor performance of the two \tfidf{-based} \plt-based methods on \mimiciii seems to be due to the fact that their \tfidf features ignore word order and are not contextualized, which is particularly important in this dataset. To confirm this, we repeated the experiments of \lwangru on \mimiciii after shuffling the words of the documents, and performance dropped by approx.\ 7.7\% across all measures, matching the performance of \plt{-based} methods.\footnote{By contrast, the drop was less significant in the other datasets (4.5\% in \eurlexdata and 2.8\% in \amazondata).} The dominance of \attentionxml in \mimiciii further supports our intuition that word order is particularly important in this dataset, as the core difference of \attentionxml with the rest of the \plt{-based} methods is the use of \rnn{-based} classifiers that use word embeddings and are sensitive to word order, instead of linear classifiers with \tfidf features, which do not capture word order. Meanwhile, in both \eurlexdata and \amazondata, the performance of \attentionxml is competitive with both \tfidf{-based} \plt{-based} methods and \lwangru, suggesting that the bag-of-words assumption holds in these cases. Thus, we can fairly assume that word order and global context (long-term dependencies) do not play a drastic role when predicting labels (concepts) on these datasets.
\vspace{1.5mm}

\noindent\textbf{Effects of transfer learning:} 
Adding context-aware \elmo embeddings to \lwangru (\lwangruelmo) improves performance across all datasets by a small margin, when considering all or frequent labels. For \eurlexdata and \amazondata, larger performance gains are obtained by fine-tuning \bertbase and \robertabase. Our proposed new method (\lwanbertbase) that employs \lwan on top of \bertbase has the best results among all methods on \eurlexdata and \amazondata, when all and frequent labels are considered. However, in both datasets, the results are comparable to \bertbase, indicating that the multi-head attention mechanism of \bert can effectively handle the large number of labels. 
\vspace{1.5mm}

\begin{table}[ht!]
\centering
\resizebox{\columnwidth}{!}{
\begin{tabular}{l|c|c|c}
  Method & $\hat{\mathrm{T}}$ & $\hat{\mathrm{F}}$& $nDCG@15$ \\
  \hline
   \attentionxml \cite{You2019} & full-text & - & \underline{\textbf{73.4}} \\
   \hline
  \bertbase \cite{bert} & 512 & 1.51 & 58.1 \\
  \robertabase \cite{roberta} & 512 & 1.45 & \underline{58.9} \\
  \hline
  \clibert \cite{alsentzer2019} & 512 & 1.60 & 58.6  \\
  \scibert \cite{beltagy2019} & 512 & 1.35 & \underline{60.5}  \\
  \hline
  \hierscibert (new) & 4096 & 1.35 & \underline{61.9}  \\
   \hline
\end{tabular}
}
\vspace*{-2mm}
\caption{Performance of \bert and its variants compared to \attentionxml on \mimiciii. $\hat{\mathrm{T}}$ is the maximum number of (possibly sub-word) tokens used per document. $\hat{\mathrm{F}}$ is the fragmentation ratio, i.e., the number of tokens (\bpe{s} or wordpieces) per word.}
\vspace*{-6mm}
\label{tab:mimicbert}
\end{table}

\begin{table*}[ht!]
{\footnotesize
\centering
\resizebox{\textwidth}{!}{
\begin{tabular}{lcc|cc|cc}
  \hline
  & \multicolumn{2}{c}{\textsc{\eurlexdata} ($K=5$)} & \multicolumn{2}{c}{\textsc{\mimiciii} ($K=15$)} & \multicolumn{2}{c}{\textsc{\amazondata} ($K=5$)} \\ 
   & \textsc{Few ($n<50)$} & \textsc{Zero} & \textsc{Few ($n<5$)} & \textsc{Zero} & \textsc{Few ($n<100$)} & \textsc{Zero} \\
  \hline
  \lwangru \cite{Chalkidis2019} & 61.7 & - & 14.3 & - & \underline{\textbf{73.6}} & - \\
  \hline
  \clwangru \cite{Rios2018-2} & 51.0 & 33.5 & 15.0 & 31.5 & 9.9 & 20.8   \\
  \dclwangru (new) & \underline{62.1} & \underline{41.5} & 19.3 & \underline{\textbf{39.3}} & \underline{39.0} & \underline{48.9}   \\
   \hline
  \dnlwangru (new) & 52.2 & 23.8 & 10.0  & 22.3 & 20.4  & 27.2  \\
  \dnclwangru (new) & \underline{62.0} & \underline{39.3}  & \underline{\textbf{23.8}} & \underline{33.6} & \underline{41.6} & \underline{47.6}   \\
   \hline
  \gclwangru \cite{Rios2018-2} & 61.8 & \underline{\textbf{42.6}}  & \underline{21.1} & \underline{35.2} & \underline{45.8} & 46.1   \\
  \gnclwangru (new)  & \underline{\textbf{62.6}} & 36.3 & 18.4 & 34.2 & 45.3 & \underline{\textbf{51.9}} \\
  \hline
\end{tabular}
}}
\caption{Results (\%) of experiments performed with zero-shot capable extensions of \lwangru. All scores are \textit{nDCG@K}, with the same \textit{K} values as in Table~\ref{tab:results}. Best results shown in bold. Best results in each zone shown underlined. $n$ is the number of training documents assigned with a label. Similar conclusions can be drawn when evaluating with \textit{RP@K} (See the appendix).}
\vspace*{-4mm}
\label{tab:variations}
\end{table*}

\noindent\textbf{Poor performance of \bert on \mimiciii:} Quite surprisingly, all three \bert{-based} models perform poorly on \mimiciii (Table~\ref{tab:results}), so we examined two possible reasons. First, we hypothesized that this poor performance is due to the distinctive writing style and terminology of biomedical documents, which are not well represented in the generic corpora these models are pre-trained on. To check this hypothesis, we employed \clibert \cite{alsentzer2019}, a version of \bertbase that has been further fine-tuned on biomedical documents, including discharge summaries. Table~\ref{tab:mimicbert} shows that \clibert performs slightly better than \bertbase on the biomedical dataset, partly confirming our hypothesis. The improvement, however, is small and \clibert still performs worse than \robertabase, which is pre-trained on larger generic corpora with a larger vocabulary. Examining the token vocabularies \cite{Gage1994} of the \bert-based models reveals that biomedical terms are frequently over-fragmented; e.g., `pneumonothorax' becomes [`p', `\#\#ne', `\#\#um', `\#\#ono', `\#\#th', `\#\#orax'], and  `schizophreniform becomes [`s', `\#\#chi', `\#\#zo', `\#\#ph', `\#\#ren', `\#\#iform']. This is also the case with \clibert, where the original vocabulary of \bertbase was retained. We suspect that such long sequences of meaningless sub-words are difficult to re-assemble into meaningful units, even when using deep pre-trained Transformer-based models. Thus we also report the performance of \scibert \cite{beltagy2019}, which was pre-trained from scratch (including building the vocabulary) on scientific articles, mostly from the biomedical domain. Indeed \scibert performs better, but still much worse than \attentionxml.

A second possible reason for the poor performance of \bert{-based} models on \mimiciii is that they can process texts only up to 512 tokens long, truncating longer documents. This is not a problem in \eurlexdata, because the first 512 tokens contain enough information to classify \eurlexdata documents (727 words on average), as shown by \citet{Chalkidis2019}. It is also not a problem in \amazondata, where texts are short (250 words on average). In \mimiciii, however, the average document length is approx.\ 1.6k  words and documents are severely truncated.\footnote{In \bpe{s}, the average document length is approx. 2.1k, as many biomedical terms are over-fragmented, thus only the 1/4 of the document actually fit in practice in \bert{-based} models.} To check the effect of text truncation, we employed a hierarchical version of \scibert, dubbed \hierscibert, similar to the hierarchical \bert of \citet{Chalkidis2019Judge}.\footnote{This model is `hierarchical' in the sense that a first layer encodes paragraphs, then another layer combines the representations of paragraphs \cite{Yang2016}. It does not use the label hierarchy.} This model encodes consecutive segments of text (each up to 512 tokens) using a shared \scibert encoder, then applies max-pooling over the segment encodings to produce a final document representation. 
\hierscibert outperforms \scibert, confirming that truncation is an important issue, but it still performs worse than \attentionxml.
We believe that a hierarchical \bert model pre-trained from scratch on biomedical corpora, especially discharge summaries, with a new \bpe vocabulary, may perform even better in future experiments.  

\subsection{Zero-shot Learning}

In Table~\ref{tab:results} we intentionally omitted zero-shot labels, as the methods discussed so far, except \gclwangru, are incapable of zero-shot learning. In general, any model that relies solely on trainable vectors to represent labels cannot cope with unseen labels, as it eventually learns to ignore unseen labels, i.e., it assigns them near-zero probabilities. In this section, we discuss the results of the zero-shot capable extensions of \lwangru (Section~\ref{sec:zero-shot)}).

In line with the experiments of \newcite{Rios2018-2},  Table~\ref{tab:variations} shows that \gclwangru (with \gcn{s}) performs better than \clwangru in zero-shot labels on all three datasets. These two zero-shot capable extensions of \lwangru also obtain better few-shot results on \mimiciii comparing to \lwangru; \gclwangru is also comparable to \lwangru in few-shot learning on \eurlexdata, but \lwangru is much better  than its two zero-shot extensions on \amazondata. The superior performance of \lwangru on \eurlexdata and \amazondata, compared to \mimiciii, is due to the fact that in the first two datasets few-shot labels are more frequent ($n \leq 50$, and $n \leq 100$, respectively) than in \mimiciii ($n \leq 5)$.\vspace{2mm}

\noindent\textbf{Are graph convolutions a key factor?} It is unclear if the gains of \gclwangru are due to the \gcn encoder of the label hierarchy, or the increased depth of \gclwangru compared to \clwangru. Table~\ref{tab:variations} shows that \dclwangru is competitive to \gclwangru, indicating that the latter benefits mostly from its increased depth, and to a smaller extent from its awareness of the label hierarchy. This motivated us to search for alternative ways to exploit the label hierarchy.\vspace{2mm}

\noindent\textbf{Alternatives in exploiting label hierarchy:}
Table~\ref{tab:variations} shows that \dnlwangru, which replaces the centroids of token embeddings of the label descriptors of \dclwangru with label embeddings produced by the \nodevec extension, is actually inferior to \dclwangru. 
In turn, this suggests that although the \nodevec extension we employed aims to encode both topological information from the hierarchy and information from the label descriptors, the centroids of word embeddings still capture information from the label descriptors that the \nodevec extension misses. This also indicates that exploiting the information from the label descriptors is probably more important than the topological information of the label hierarchy for few and zero-shot learning generalization.

\dnclwangru, which combines the centroids with the label embeddings of the \nodevec extension, is comparable to \dclwangru, while being better overall in few-shot labels.
Combining the \gcn encoder and the \nodevec extension (\gnclwangru) leads to a large improvement in zero-shot labels (46.1\% to 51.9\% \textit{nDCG@K}) on \amazondata. On \eurlexdata, however, the original \gclwangru still has the best zero-shot results; and on \mimiciii, the best zero-shot results are obtained by the hierarchy-unaware \dclwangru. These mixed findings seem related to the \gap of each dataset (Fig.~\ref{fig:hierarchies}).\vspace{2mm}

\noindent\textbf{The role of graph-aware annotation proximity:} When gold label assignments are dense, neighbouring labels co-occur more frequently, thus models can leverage topological information and learn how to better cope with neighbouring labels, which is what both \gcn{s} and \nodevec do. The denser the gold label assignments, the more we can rely on more distant neighbours, and the better it becomes to include graph embedding methods that conflate larger neighbourhoods, like \nodevec (included in \gnclwangru) on \amazondata (\gap 0.86), when predicting unseen labels. 

For medium proximity gold label assignments, as in \eurlexdata (\gap 0.45), it seems preferable to rely on closer neighbours only; hence, it is better to use only graph encoders that conflate smaller neighbourhoods, like the \gcn{s} which apply convolution filters to neighbours up to two hops away, as in \gclwangru (excl. \nodevec extension). 

When label assignments are sparse, as in \mimiciii (\gap 0.27), where only non-neighbouring leaf labels are assigned in the same document, leveraging the topological information (e.g., knowing that a rare label shares an ancestor with a frequent one) is not always helpful, which is why encoding the label hierarchy shows no advantage in zero-shot learning in \mimiciii; however, it can still be useful when we at least have few training instances, as the few-shot results of \mimiciii indicate.\vspace{2mm}

\noindent Overall, we conclude that the \gcn label hierarchy encoder does not always improve \lwan{s} in zero-shot learning, compared to equally deep \lwan{s}, and that depending on the proximity of label assignments (based on the label annotation guidelines) it may be preferable to use additional or no hierarchy-aware encodings for zero-shot learning.

\vspace*{-1mm}

\section{Conclusions}

We presented an extensive study of \lmtc methods in three domains, to answer three understudied questions on (1) the competitiveness of \plt-based methods against neural models, (2) the use of the label hierarchy, (3) the benefits from transfer learning. A condensed summary of our findings is that (1) \tfidf \plt-based methods are definitely worth considering, but are not always competitive, while \attentionxml, a neural \plt{-based} method that captures word order, is robust across datasets; (2) transfer learning leads to state-of-the-art results in general, but \bert-based models can fail spectacularly when documents are long and technical terms get over-fragmented; (3) the best way to use the label hierarchy in neural methods depends on the proximity of the label assignments in each dataset. An even shorter summary is that no single method is best across all domains and label groups (all, few, zero) as the language, the size of documents, and the label assignment strongly vary with direct implications in the performance of each method. 

In future work, we would like to further investigate few and zero-shot learning in \lmtc, especially in \bert models that are currently unable to cope with zero-shot labels. It is also important to shed more light on the poor performance of \bert models in \mimiciii and propose alternatives that can cope both with long documents \cite{kitaev2020,beltagy2020} and domain-specific terminology, reducing word over-fragmentation. Pre-training \bert from scratch on discharge summaries with a new \bpe vocabulary is a possible solution. Finally, we would like to combine \plt{s} with \bert, similarly to \attentionxml, but the computational cost of fine-tuning multiple \bert encoders, one for each \plt node, would be massive, surpassing the training cost of very large Transformer-based models, like \tfive~\citep{2019t5} and \megatronlm~\cite{2019megatronlm} with billions of parameters (30-100x the size of \bertbase).

\bibliographystyle{acl_natbib}
\bibliography{emnlp2020}

\appendix

\section{Additional Implementation Details}
All experiments were deployed in \textsc{nvidia gt1080ti} \textsc{gpu} cards, in a single \textsc{gpu} fashion. In Table~\ref{tab:params}, we report the size of the models and the elapsed training time. Hyper-parameters were tuned using \textsc{hyperopt},\footnote{\url{https://github.com/hyperopt/hyperopt}} selecting  values with the best loss on development data. Table~\ref{tab:parameters} shows the hyper-parameters search space and the selected values. We use 200-D pretrained \glove embeddings \cite{pennington2014glove} for \eurlexdata and \amazondata, and 200-D \wordvec embeddings pretrained on \textsc{pubmed}\footnote{\url{https://www.ncbi.nlm.nih.gov/pubmed/}} \citep{McDonald2018} for \mimiciii. For \bert-based methods we tuned only the learning rate, considering the values \{$2e\textrm{-}5$, $3e\textrm{-}5$, $5e\textrm{-}5$\}, selecting $2e\textrm{-}5$ for \eurlexdata and \amazondata, and $5e\textrm{-}5$ for \mimiciii. Finally, for \parabel and \bonsai we tuned the $n$-gram order in the range \{1, 2, 3, 4, 5\}, and the number of $n$-gram features in the range \{100k, 200k, 300k, 400k\}. When $n>1$ we use $n$-grams up to order $n$, e.g. for $n=3$ we use 1-grams, 2-grams and 3-grams. In all datasets the optimal values were 200k features for $n=5$.

\begin{table*}[t!]
{\footnotesize
\centering
\resizebox{\textwidth}{!}{
\begin{tabular}{lccccc}
  \hline
  \hline
  & \multicolumn{5}{c}{\eurlexdata} \\
  \hline
  \hline
  Search space & Layers & Units & Dropout & Word Dropout & Batch Size \\
  \hline
  \textsc{Baselines} & [1, 2] & [100, 200, 300, 400] & [0.1, 0.2, 0.3] & [0, 0.01, 0.02] & [8, 16] \\
  \hline
  \lwangru \cite{Chalkidis2019}  & 1 & 300 & 0.4 & 0 & 16 \\
  \hline
  \textsc{Zero-shot} & [1, 2] & [100, 200, 300, 400] & [0.1, 0.2, 0.3] & [0, 0.01, 0.02] & [8, 16] \\
  \hline
  \clwangru \cite{Rios2018-2} & 1 & 100 & 0.1 & 0.02 & 16 \\
  \gclwangru \cite{Rios2018-2}  & 1 & 100 & 0.1 & 0 & 16 \\
  \dclwangru (new) & 1 & 100 & 0.1 & 0 & 16 \\
  \dnlwangru (new) & 1 & 100 & 0.1 & 0 & 16 \\
  \dnclwangru (new) & 1 & 100 & 0.1 & 0 & 16 \\
  \gnclwangru (new) & 1 & 100 & 0.1 & 0.02 & 16\\
  \hline
  \textsc{Transfer learning} & [12] & [768] & [0.1, 0.2, 0.3] & - & [8, 16] \\
  \hline
  \bertbase \cite{bert} & 12 & 768 & 0.1 & - & 8 \\
  \robertabase \cite{roberta} & 12 & 768 & 0.1 & - & 8 \\
  \lwanbert (new) & 12 & 768 & 0.1 & - & 8 \\
  \hline
  \hline
  & \multicolumn{5}{c}{\mimiciii} \\
  \hline
  \hline
  Search space & Layers & Units & Dropout & Word Dropout & Batch Size \\
  \hline
  \textsc{Baselines} & [1, 2] & [100, 200, 300, 400] & [0.1, 0.2, 0.3] & [0, 0.01, 0.02] & [8, 16] \\
  \hline
  \lwangru \cite{Chalkidis2019} & 2 & 300 & 0.3 & 0 & 8 \\
  \hline
  \textsc{Zero-shot} & [1, 2] & [100, 200, 300, 400] & [0.1, 0.2, 0.3] & [0, 0.01, 0.02] & [8, 16] \\
  \hline
  \clwangru \cite{Rios2018-2} & 2 & 100 & 0.1 & 0 & 8 \\
  \gclwangru \cite{Rios2018-2} & 1 & 100 & 0.1 & 0 & 8 \\
  \dclwangru (new) & 1 & 100 & 0.1 & 0 & 8 \\
  \dnlwangru (new) & 1 & 100 & 0.1 & 0 & 8 \\
  \dnclwangru (new) & 1 & 100 & 0.1 & 0 & 8 \\
  \gnclwangru (new) & 1 & 100 & 0.1 & 0 & 8 \\
  \hline
  \textsc{Transfer learning} & [12] & [768] & [0.1, 0.2, 0.3] & - & [8, 16] \\
  \hline
  \bertbase \cite{bert} & 12 & 768 & 0.1 & - & 8 \\
  \robertabase \cite{roberta} & 12 & 768 & 0.1 & - & 8 \\
  \lwanbert (new) & 12 & 768 & 0.1 & - & 8 \\
  \hline
  \hline
  & \multicolumn{5}{c}{\amazon} \\
  \hline
  \hline
  Search space & Layers & Units & Dropout & Word Dropout & Batch Size \\

  \hline
  \textsc{Baselines} & [1, 2] & [100, 200, 300, 400] & [0.1, 0.2, 0.3] & [0, 0.01, 0.02] & [8, 16] \\
  \hline
  \lwangru \cite{Chalkidis2019} & 2 & 300 & 0.1 & 0 & 32 \\
  \hline
  \textsc{Zero-shot} & [1, 2] & [100, 200, 300, 400] & [0.1, 0.2, 0.3] & [0, 0.01, 0.02] & [8, 16] \\
  \hline
  \clwangru \cite{Rios2018-2} & 2 & 100 & 0.1 & 0 & 32 \\
  \gclwangru \cite{Rios2018-2} & 1 & 100 & 0.1 & 0 & 32 \\
  \dclwangru (new) & 2 & 100 & 0.1 & 0 & 32 \\
  \dnlwangru (new) & 1 & 100 & 0.1 & 0 & 32 \\
  \dnclwangru (new) & 2 & 100 & 0.1 & 0 & 32 \\
  \gnclwangru (new) & 1 & 100 & 0.1 & 0 & 32 \\
  \hline
  \textsc{Transfer learning} & [12] & [768] & [0.1, 0.2, 0.3] & - & [8, 16] \\
  \hline
  \bertbase \cite{bert} & 12 & 768 & 0.1 & - & 8 \\
  \robertabase \cite{roberta} & 12 & 768 & 0.1 & - & 8 \\
  \lwanbert (ours) & 12 & 768 & 0.1 & - & 8 \\
  \hline
 \end{tabular}
 }}
\caption{Hyper-parameter search space and best values chosen for all neural methods except \bert-based ones.}
\label{tab:parameters}
\vspace*{35mm}
\end{table*}

\section{BIGRUs vs.\ CNNs}
\citet{Chalkidis2019} showed that \bigru{s} are better encoders than \cnn{s} in \eurlexdata. We confirm these findings across all datasets (Table~\ref{tab:cnns-grus}). \lwangru, \clwangru and \gclwangru outperform \lwancnn, \clwancnn and \gclwancnn by 3.5 to 16.5 percentage points.

\begin{table*}[t!]
{\footnotesize
\centering
\resizebox{\textwidth}{!}{
\begin{tabular}{lcc|cc|cc|cc}
  \hline
  & \multicolumn{2}{c}{\textsc{All Labels}} & \multicolumn{2}{c}{\textsc{Frequent}} & \multicolumn{2}{c}{\textsc{Few}} & \multicolumn{2}{c}{\textsc{Zero}} \\ 
  & $RP@K$ & $nDCG@K$ & $RP@K$ & $nDCG@K$ & $RP@K$ & $nDCG@K$ & $RP@K$ & $nDCG@K$\\
  \hline
  \hline
  & \multicolumn{7}{c}{\eurlexdata ($L_{AVG}=5.07, K=5$)}  & \\
  \hline
  \hline
  \lwangru & \textbf{77.1} & \textbf{80.1} & \textbf{81.0} & \textbf{82.4} & \textbf{65.6} & \textbf{61.7} & - & - \\ 
  \lwancnn & 71.7 & 74.6 & 76.1 & 77.3 & 61.1 & 55.1 & - & - \\
  \hline
  \clwangru & \textbf{72.0} & \textbf{75.6} & \textbf{76.9} & \textbf{78.7} & \textbf{55.7} & \textbf{51.0} & \textbf{46.1} & \textbf{33.5} \\
  \clwancnn & 68.5 & 71.7 & 73.2 & 74.5 & 49.7 & 45.7 & 36.1 & 29.9 \\
  \hline
  \gclwangru & \textbf{76.8} & \textbf{80.0} & \textbf{80.6} & \textbf{82.3} & \textbf{66.2} & \textbf{61.8} & \textbf{48.9} & \textbf{42.6} \\
  \gclwancnn & 70.9 & 74.4 & 75.4 & 77.2 & 52.3 & 48.4 & 37.1 & 29.6 \\
  \hline
  \hline
  & \multicolumn{7}{c}{\mimiciii ($L_{AVG}=15.45, K=15$)}  & \\
  \hline
  \hline
  \lwangru & \textbf{66.2} & \textbf{70.1} & \textbf{66.8} & \textbf{70.6} & \textbf{21.7} & \textbf{14.3} & - & - \\
  \lwancnn & 60.5 & 64.3 & 61.1 & 64.7 & 16.3 & 10.2 & - & - \\
  \hline
  \clwangru & \textbf{60.2} & \textbf{64.9} & \textbf{60.9} & \textbf{65.3} & \textbf{26.9} & \textbf{15.0} & \textbf{52.6} & \textbf{31.5} \\
  \clwancnn & 54.9 & 59.5 & 55.5 & 59.9 & 21.2 & 11.7 & 37.3 & 19.5 \\
  \hline
  \gclwangru & \textbf{64.9} & \textbf{69.1} & \textbf{65.6} & \textbf{69.6} & \textbf{35.9} & \textbf{21.1} & \textbf{56.6} & \textbf{35.2} \\
  \gclwancnn & 56.6 & 60.9 & 57.2 & 61.3 & 23.7 & 13.0 & 38.2 & 22.2 \\
  \hline
  \hline
  & \multicolumn{7}{c}{\amazondata ($L_{AVG}=5.04, K=5$)}  & \\
  \hline
  \hline
  \lwangru & \textbf{83.9} & \textbf{85.4} & \textbf{84.9} & \textbf{86.1} & \textbf{80.0} & \textbf{73.6} & - & - \\
  \lwancnn & 77.1 & 79.1 & 78.2 & 79.7 & 70.4 & 63.6 & - & - \\
  \hline
  \clwangru & \textbf{64.6} & \textbf{68.2} & \textbf{67.2} & \textbf{70.3} & \textbf{13.8} & \textbf{9.9} & \textbf{29.9} & \textbf{20.8} \\
  \clwancnn & 56.2 & 59.2 & 58.6 & 61.2 & 8.6 & 6.3 & 19.5 & 14.5 \\
  \hline
  \gclwangru & \textbf{77.4} & \textbf{79.8} & \textbf{79.1} & \textbf{81.0} & \textbf{53.7} & \textbf{45.8} & \textbf{56.1} & \textbf{46.1} \\
  \gclwancnn & 72.6 & 75.3 & 74.3 & 76.4 & 41.3 & 34.0 & 45.6 & 34.5 \\
  \hline
\end{tabular}
}}
\caption{Results (\%) of experiments performed to compare \gru vs.\ \cnn encoders. 
Best results in each zone shown in bold. We show results for $K$ close to the average number of labels $L_{AVG}$.}
\label{tab:cnns-grus}
\end{table*}

\begin{table*}[t!]
{\footnotesize
\centering
\resizebox{\textwidth}{!}{
\begin{tabular}{lccc}
  \hline
  Methods & Parameters & Trainable Parameter & Train Time \\
  \hline
  \multicolumn{4}{c}{\textsc{Baselines}}  \\
  \hline
  \lwangru \cite{Chalkidis2019}  & 86 & 6 & 14h \\
  \hline
  \multicolumn{4}{c}{\textsc{Zero-shot}} \\
  \hline
  \clwangru \cite{Rios2018-2} & 80.2 & 0.2 & 9.3h \\
  \gclwangru \cite{Rios2018-2}  & 80.5 & 0.5 & 18.5h \\
  \dclwangru (new) & 81.3 & 1.3 & 11.2h \\
  \dnlwangru (new) & 80.2 & 0.2 & 9.5h \\
  \dnclwangru (new) & 81.6 & 1.6 & 10.1h \\
  \gnclwangru (new) & 80.5 & 0.5 & 20.2h \\
  \hline
  \multicolumn{4}{c}{\textsc{Transfer learning}} \\
  \hline
  \bertbase \cite{bert} & 110 & 110 & 9.5h  \\
  \robertabase \cite{roberta} & 110 & 110 & 9.5h \\
  \lwanbert (new) & 119 & 119 & 11h \\
  \hline
  \hline
\end{tabular}
}}
\caption{Number of parameters (trainable or not) in millions and training time for a single run reported for all examined methods.}
\label{tab:params}
\end{table*}

\section{Additional Results}

Table~\ref{tab:variations_app} shows \textit{RP@K} results of the zero-shot capable methods. As with \textit{nDCG@K}, we conclude that the \gcn label hierarchy encoder of \citet{Rios2018-2} does not always improve \lwan{s} in zero-shot learning, compared to equally deep \lwan{s}, and that depending on the proximity of label assignments, it may be preferable to use additional or no encodings of the hierarchy for zero-shot learning. Also, the zero-shot capable methods outperform \lwangru in all, frequent, and few labels, but no method is consistently the best.

\begin{table*}[t!]
{\footnotesize
\centering
\resizebox{\textwidth}{!}{
\begin{tabular}{lcc|cc|cc}
  \hline
  & \multicolumn{2}{c}{\textsc{\eurlexdata} ($K=5$)} & \multicolumn{2}{c}{\textsc{\mimiciii} ($K=15$)} & \multicolumn{2}{c}{\textsc{\amazondata} ($K=5$)} \\\ 
   & \textsc{Few ($n<50)$} & \textsc{Zero} & \textsc{Few ($n<5$)} & \textsc{Zero} & \textsc{Few ($n<100$)} & \textsc{Zero} \\ 
  \hline
  \lwangru \cite{Chalkidis2019} & 65.6 & - & 21.7 & - & 80.0 & - \\
  \hline
  \clwangru \cite{Rios2018-2} & 55.7 & 46.1 & 26.9 & 52.6 & 13.8 & 29.9   \\
  \dclwangru (new) & \underline{66.8} & \underline{\textbf{53.9}} & 33.6 & \underline{\textbf{63.9}} & \underline{47.0} & \underline{57.1}   \\
   \hline
  \dnlwangru (new) & 56.9 & 34.3 & 19.5  & 43.9 & 27.1  & 36.9  \\
  \dnclwangru (new) & \underline{66.9} & \underline{51.7}  & \underline{\textbf{41.3}} & \underline{59.4} & \underline{50.2} & \underline{59.6}   \\
   \hline
  \gclwangru \cite{Rios2018-2} & 66.2 & 48.9  & \underline{35.9} & 56.6 & 53.7 & 56.1   \\
  \gnclwangru (new)  & \underline{\textbf{67.7}} & \underline{49.4} & 31.6 & \underline{57.5} & \underline{\textbf{53.8}} & \underline{\textbf{63.4}} \\
  \hline
\end{tabular}
}}
\caption{Results (\%) of experiments performed with zero-shot capable extensions of \lwangru. All scores are \textit{RP@K}, with the same \textit{K} values as in Table 1 of the main paper. Best results of zero-shot capable methods (excluding \lwangru) shown in bold. Best results in each zone shown underlined. $n$ is the number of training documents assigned with a label.}
\vspace*{-4mm}
\label{tab:variations_app}
\end{table*}

\end{document}